\def\year{2022}\relax
\documentclass[letterpaper]{article}
\usepackage{aaai22} 
\usepackage{times} 
\usepackage{helvet} 
\usepackage{courier} 
\usepackage[hyphens]{url} 
\usepackage{graphicx} 
\usepackage{graphicx}  
\usepackage{natbib}  
\usepackage{caption}  
\DeclareCaptionStyle{ruled}%
  {labelfont=normalfont,labelsep=colon,strut=off}
\usepackage{amsmath}
\usepackage{amssymb}
\usepackage{multicol}
\usepackage{multirow}
\usepackage{subcaption}
\usepackage{makecell}

\DeclareMathOperator*{\argmax}{arg\,max}

\usepackage{algorithm}
\usepackage{algorithmic}
\usepackage{pifont}
\newcommand{\cmark}{\ding{51}}%
\newcommand{\xmark}{\ding{55}}%

\usepackage{xcolor,colortbl}
\definecolor{LightCyan}{rgb}{0.88,1,1}

\usepackage{newfloat}
\usepackage{listings}
\floatstyle{ruled}
\newfloat{listing}{tb}{lst}{}
\floatname{listing}{Listing}

\usepackage{xcolor}

\title{SASA: Semantics-Augmented Set Abstraction for Point-based 3D Object Detection}
\author{
    Chen Chen\textsuperscript{\rm 1},
    Zhe Chen\textsuperscript{\rm 1},
    Jing Zhang\textsuperscript{\rm 1},
    Dacheng Tao\textsuperscript{\rm 2,\rm 1}
}
\affiliations{
    \textsuperscript{\rm 1} The University of Sydney, Australia\quad
    \textsuperscript{\rm 2} JD Explore Academy, China\\
    cche9000@uni.sydney.edu.au, \{zhe.chen1, jing.zhang1\}@sydney.edu.au, dacheng.tao@gmail.com
}

\begin{document}

\maketitle

\begin{abstract}
Although point-based networks are demonstrated to be accurate for 3D point cloud modeling, they are still falling behind their voxel-based competitors in 3D detection. We observe that the prevailing set abstraction design for down-sampling points may maintain too much unimportant background information that can affect feature learning for detecting objects. To tackle this issue, we propose a novel set abstraction method named Semantics-Augmented Set Abstraction (SASA). Technically, we first add a binary segmentation module as the side output to help identify foreground points. Based on the estimated point-wise foreground scores, we then propose a semantics-guided point sampling algorithm to help retain more important foreground points during down-sampling. In practice, SASA shows to be effective in identifying valuable points related to foreground objects and improving feature learning for point-based 3D detection. Additionally, it is an easy-to-plug-in module and able to boost various point-based detectors, including single-stage and two-stage ones. Extensive experiments on the popular KITTI and nuScenes datasets validate the superiority of SASA, lifting point-based detection models to reach comparable performance to state-of-the-art voxel-based methods. Code will be available at \url{https://github.com/blakechen97/SASA}.
\end{abstract}

\section{Introduction}
3D object detection has attracted increasing interest from researchers because it plays an important role in various real-world scenarios like autonomous driving and the robotic system \cite{shi2020pv, yang20203dssd}. 
This task aims to identify and localize objects from 3D scenes. To properly detect objects from 3D space, LiDAR sensors are widely applied to capture 3D point clouds and represent the surrounding environments. Comparing to RGB images, point clouds provide rich and accurate 3D structure information, which is important for precise 3D object localization.

\begin{figure}[t]
\centering
\includegraphics[width=\linewidth]{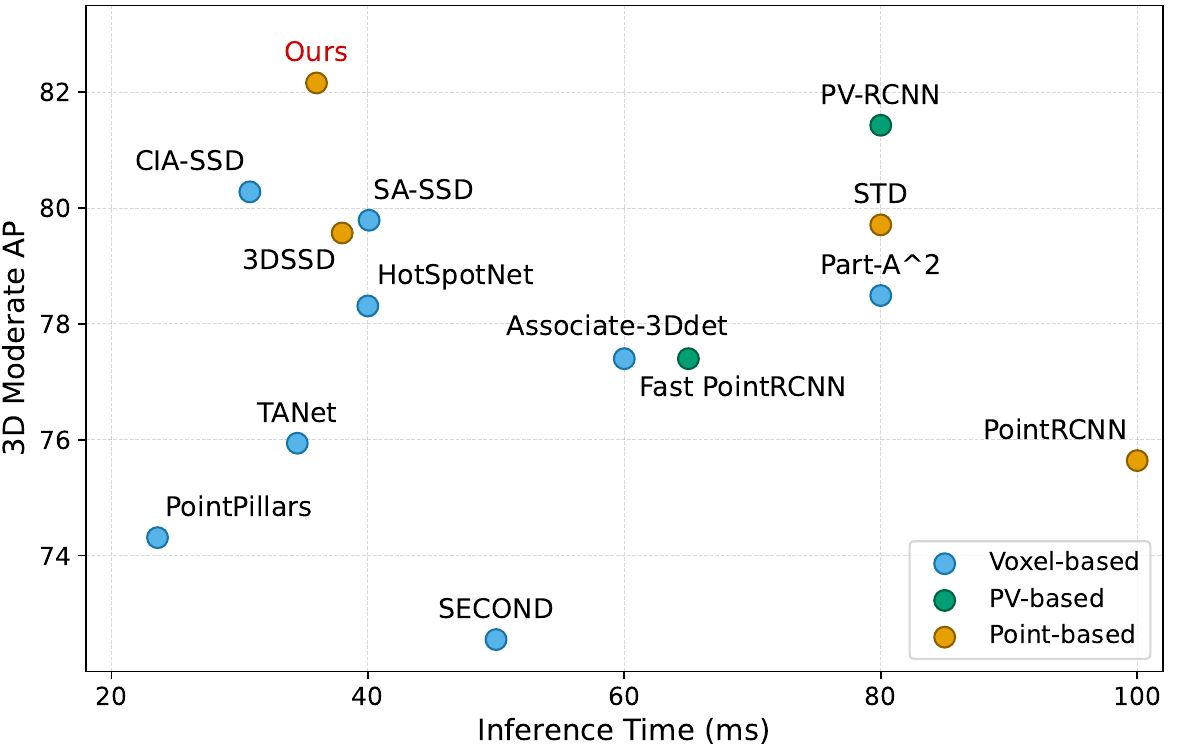}
\caption{Our method reaches top performance (moderate AP: $82.16\%$) among both voxel-based and point-based detectors with a high inference speed for the car detection on the KITTI benchmark \cite{geiger2012we}.}
\label{fig_intro}
\end{figure}

To exploit the representational power of deep learning \cite{zhang2020empowering}, researchers have designed different neural networks to extract 3D features, including voxel-based ones \cite{zhou2018voxelnet, yan2018second, deng2020voxel} that discretize sparse points into regular voxel grids and point-based ones \cite{shi2019pointrcnn, qi2019deep, yang20203dssd} that directly perform feature learning on 3D points. Benefiting from transformation-free point cloud processing and flexible receptive fields \cite{shi2020pv}, point-based methods have the potential of achieving compelling performance \cite{yang20203dssd}. However, compared with voxel-based detectors that show vast development, point-based 3D detection stagnates in recent years and fails to achieve top performance on related datasets.

By investigating popular point-based methods, we find that an important problem is that the widely-used \textit{set abstraction} (SA) is inefficient to describe scenes in the context of detection, especially with the problematic sampling strategy. In particular, the SA layer first selects a subset of input points as key points and then encodes context representations from nearby points for each sampled key point. However, when selecting key points, existing sampling strategies, which tend to choose distant points to better cover the entire scene, could make abstracted point sets involve excessive irrelevant background points like points on the ground, since the majority of the 3D space belongs to background especially in outdoor scenarios \cite{chen2019progressive}. These irrelevant points usually deliver trivial information for detecting objects and, at the same time, plenty of beneficial foreground points could be inappropriately discarded. For example, points on small objects like pedestrians may be completely neglected. Consequently, the point set given by SA may fail to provide sufficient foreground information or cover many foreground instances, thus the detection performance could be largely degraded. Although most point-based detectors \cite{qi2018frustum, yang2019std, shi2019pointrcnn} apply \textit{feature propagation} (FP) layers to retrieve the foreground points dropped in the previous SA stage, these FP layers bring heavy memory usage and high computational cost \cite{yang20203dssd} inevitably.

To solve the issue, we propose a \emph{Semantics-Augmented Set Abstraction} (SASA) for point-based 3D detection. By incorporating point-wise semantic cues, we can help avoid including too many potentially irrelevant background points and focus on more informative foreground ones in the SA stage. Hence, abstracted point sets could then provide more object-related information for the succeeding box prediction network. To properly incorporate point semantics into SA, we have made the following two updates to the SA layer in Pointnet++ \cite{qi2017pointnet++}. Firstly, we add a point binary segmentation module to identify foreground points from the input. Then, given point semantic maps, we adopt a novel sampling algorithm, \emph{semantics-guided farthest point sampling} (S-FPS), to choose representative key points for SA layers. Comparing to the commonly used \emph{farthest point sampling} (FPS), our proposed S-FPS gives more preference to positive points so more points from foreground are kept through down-sampling. With point-wise segmentation and advanced sampling strategy, SASA serves as a strong point feature learning technique for 3D detection.

In practice, our proposed SASA is an easy-to-plug-in module and can work seamlessly with various point-based detection frameworks. We have successfully implemented it in two popular point-based baselines, 3DSSD \cite{yang20203dssd} and PointRCNN \cite{shi2019pointrcnn}. Though they use way different feature learning and box prediction schemes, SASA delivers consistent improvement. Experimental results (Sec. \ref{sec:experiments}) show that SASA can boost the mean average precision (mAP) by around $2\%$ for the most competitive car class on the KITTI dataset \cite{geiger2012we} and show notable improvement on the large-scale nuScenes dataset \cite{caesar2020nuscenes}.

In summary, the contribution of this work is derived from our novel point set abstraction design with semantics. For point-based 3D detection, we (a) attach a binary segmentation module to the SA layer to identify valuable points from foreground and; (b) propose a novel sampling algorithm S-FPS to make abstracted point sets focus on object areas. Our design is lightweight and can be easily adopted in manifold point-based detection models. Experimental results show that our method obtains highly boosted results on both single-stage and two-stage baselines on the KITTI \cite{geiger2012we} and nuScenes \cite{caesar2020nuscenes} datasets and sets new state-of-the-art for point-based 3D object detection.

\section{Related Work}

\subsubsection{3D Object Detection from Point Clouds.}
According to the 3D point processing schemes, recent 3D detection models can be mainly divided into grid-based and point-based methods. Grid-based methods \cite{chen2017multi, ku2018joint, song2016deep, zhou2018voxelnet, yan2018second, chen2019fast, lang2019pointpillars, he2020structure, shi2020points, deng2020voxel} firstly transform unordered 3D points into regular 2D pixels or 3D voxels where convolutional neural networks (CNN) can be applied for point cloud modeling. Some methods \cite{beltran2018birdnet, lang2019pointpillars} process point clouds from projected 2D views (\emph{e.g.}  bird's eye view). VoxelNet \cite{zhou2018voxelnet} proposes to model 3D scenes via voxelization and 3D CNN. SECOND \cite{yan2018second} formulates an elegant 3D feature learning backbone with sparse convolutions \cite{liu2015sparse} and makes a fast and effective one-stage detector. VoxelRCNN \cite{deng2020voxel} proposes a novel voxel RoI pooling to efficiently aggregate RoI features from voxels in a Pointnet \cite{qi2017pointnet++} set abstraction style.

Another stream is point-based detection. Based on the prevailing point feature learning technique, Pointnet \cite{qi2017pointnet, qi2017pointnet++}, these methods model point clouds from raw points input. F-Pointnet \cite{qi2018frustum} firstly introduces Pointnet \cite{qi2017pointnet, qi2017pointnet++} to 3D detection for locating objects from cropped point clouds given by 2D detectors. To avoid leveraging RGB images, PointRCNN \cite{shi2019pointrcnn} proposes a fully point-based detection paradigm, comprising a point-based region proposal network (RPN) to generate 3D proposals from point-wise features and a point-based refinement network to adjust 3D boxes with internal point features. VoteNet \cite{qi2019deep} replaces point-based RPN with a lightweight voting scheme and obtains an anchor-free point-based detector. 3DSSD \cite{yang20203dssd} adopts a more advanced point sampling strategy to safely remove expensive FP layers without hurting the detection recall. Based on these popular point-based detection frameworks, we further explore how to upgrade the fundamental feature learning phase for boosting point-based detection.

\subsubsection{Sampling Algorithms for Set Abstraction.}
In Pointnet-based feature learning paradigms \cite{qi2017pointnet++}, SA layers firstly sample a subset of input points for dimension reduction, where most point-based models \cite{qi2018frustum, shi2019pointrcnn, qi2019deep} adopt the classic \textit{farthest point sampling} (FPS) algorithm for key points sampling. Recent works \cite{yang2019modeling, lang2020samplenet, yang20203dssd, nezhadarya2020adaptive} devise novel sampling algorithms to obtain better point modeling ability. For the representative point cloud classification task, \cite{yang2019modeling, lang2020samplenet} manage to make the sampling process differentiable so it can be optimized in an end-to-end manner. Besides, some methods choose to involve additional heuristic information into the sampling strategy. For example, \citet{nezhadarya2020adaptive} tends to keep critical points that occupy a large proportion of channels in final representations. In 3D object detection, \citet{yang20203dssd} proposes the Feature-FPS (F-FPS) where the feature distance between points is also considered to increase the feature diversity of sampled points. In this paper, we use a more direct heuristic cue, point semantics, to help SA layers focus on more beneficial points from foreground.

\begin{figure*}
\centering
\includegraphics[width=\linewidth]{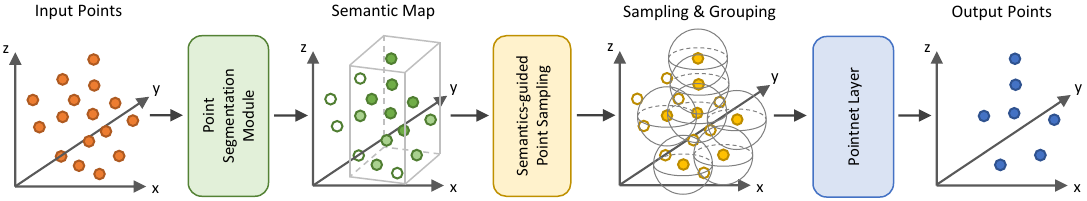}
\caption{The structure of our proposed Semantics-Augmented Set Abstraction (SASA) layer. Based on the original SA layer design, we add a point segmentation module for mapping input point features to binary segmentation masks and update the point sampling algorithm by our semantics-guided farthest point sampling (S-FPS). Point semantic labels are derived from ground-truth boxes and all point segmentation modules are optimized with a segmentation loss function in an end-to-end manner.}
\label{fig:architecture}
\end{figure*}

\section{Semantics-Augmented Set Abstraction}

The overall structure of SASA is shown in Figure \ref{fig:architecture}, which mainly comprises three components: a point segmentation module, a semantics-guided point sampling layer and a normal Pointnet++ SA layer.

Given the input point coordinates $\mathbf{X}$ and features $\mathbf{F}$, we first feed the point features to the point segmentation module to compute the point-wise foreground scores $\mathbf{P}$. Then, we employ our S-FPS to sample key point set $\mathbf{K}$ based on point coordinates $\mathbf{X}$ and foreground scores $\mathbf{P}$. For each point in the key point set $\mathbf{K}$, we apply the normal Pointnet++ SA layer \cite{qi2017pointnet++}, including a point grouping operation, a multi-layer perceptron (MLP) and a max-pooling layer, to calculate high-level representations for sampled key points. The output key point coordinates and features are sent to the succeeding networks for further processing.

\subsection{Point Segmentation Module}

To help the Pointnet build the awareness of local semantics, we embed a lightweight point segmentation module in SASA. It is a simple $2$-layer MLP and classifies input points into two categories, \textit{i.e.} foreground and background. Specifically, denoting $\{f^{(l_k)}_1,f^{(l_k)}_2,\dots,f^{(l_k)}_{N_k} \}$ as the $l_k$-dimension point features fed to the $k$-th SA layer, the foreground score $p\in[0, 1]$ for each point is calculated as:
\begin{equation}
  p_i = \sigma(\mathcal{M}_k(f^{(l_k)}_i))~,
  \label{eq:point_cls_layer}
\end{equation}
where $\mathcal{M}_k(\cdot)$ denotes the point segmentation module within the $k$-th SA layer, mapping input point-wise features $f_i$ to foreground scores $p_i$. $\sigma(\cdot)$ is the sigmoid function.

For training the point segmentation module in each SASA layer, foreground segmentation labels for points can be naturally derived from box annotations. Similar to \cite{shi2019pointrcnn}, points inside any one of the ground-truth 3D bounding boxes are regarded as foreground points and the others as background ones. The total segmentation loss is computed with a cross entropy (CE) loss function:
\begin{equation}
    \mathcal{L}_{seg} = \sum_{k=1}^{m}{
        \frac{\lambda_k}{N_k} \cdot 
        \sum_{i=1}^{N_k}{
            \text{CE}(p_i^{[k]}, \Hat{p}_i^{[k]})
        }
    },
\label{eq:seg}
\end{equation}
where $p_i^{[k]}$ and $\Hat{p}_i^{[k]}$ denote the predicted foreground score and the ground-truth segmentation label ($1$ for points from foreground and $0$ for ones from background) of the $i$-th point in the $k$-th SA layer. $N_k$ and $\lambda_k$ are the total number of input points and the segmentation loss weight for the $k$-th SA layer. The detailed parameter setting is deferred to Sec. \ref{sec:implementation_details}.

\subsection{Semantics-guided Farthest Point Sampling}
Local semantic perception indicates hotspot regions where objects of interest may exist. Considering the goal of detecting objects, we need to pay more attention to these locations and sample more points from there. To exploit obtained point semantics in the sampling stage, a straightforward way could be directly choosing points with top-K foreground scores, but we have observed that this method selects too many points from easily identified objects, which usually have much higher foreground scores. The obtained key point set fails to cover the 3D scene and a great proportion of ground-truth objects are ignored. Thus, the overall detection performance is largely hurt.

Hence, we propose a novel point sampling algorithm, \textit{i.e.} semantics-guided farthest point sampling (S-FPS), for incorporating the global scene awareness of FPS and the local object awareness induced by semantic heuristics. Given point-wise semantics yielded by the previous segmentation module as well as point coordinates from input, the process of our proposed S-FPS is described in Algorithm \ref{alg:S-FPS}. The main idea is to select more foreground points by giving precedence to the points with higher foreground scores. Remaining the overall procedure of FPS unchanged, we rectify the sampling metric, distance to already-sampled points, with point foreground scores. Specifically, given 3D coordinates $\{x_1,x_2,\dots,x_N\}$ and foreground scores $\{p_1,p_2,\dots,p_N\}$ of input points, a distance array $\{d_1,d_2,\dots,d_N\}$ maintains the shortest distance from $i$-th point to already selected key points. In each round of selection, we add the point with highest \emph{semantics-weighted distance} $\Tilde{d}_i$ to the key point set and it is computed as:
\begin{equation}
    \Tilde{d}_i = p_i^\gamma \cdot d_i ~,
\end{equation}
where $\gamma$ is the balance factor controlling the importance of semantic information. It is worth noticing that S-FPS can reduce to vanilla FPS when $\gamma=0$ and can also approximate to the aforementioned top-K selection if $\gamma$ becomes extremely large.

The benefit brought by S-FPS is manifold. Firstly, S-FPS can retain diverse points from foreground. Incorporated with semantic weights, positive points are more favored than negative ones during sampling, since they usually have a larger semantics-guided distance. Secondly, S-FPS enhances the density of key points in high-score areas, where foreground objects exist with higher probabilities. This could help provide more beneficial information for the follow-up box prediction network. Also, S-FPS is less sensitive to distant outliers \cite{yang2019modeling}. Though outliers usually have a larger distance to other points, their low semantic weights prevent the sampling algorithm from choosing them. Lastly, S-FPS is permutation-irrelevant \cite{yang2019modeling}. That is, previous sampling algorithms like FPS and F-FPS do not have a specific start point so different orders of input points may lead to different sampling outcomes. While S-FPS always starts with the point with the highest semantic score and all succeeding selections are unique. Key points sampled by S-FPS remain stable against different permutations.

\begin{algorithm}[t]
    \caption{Semantics-guided Farthest Point Sampling Algorithm. $N$ is the number of input points and $M$ is the number of output points sampled by the algorithm.}
    \label{alg:S-FPS}
    \begin{tabular}{l@{}l}
        \textbf{Input: } & coordinates $\mathbf{X}=\{x_1,\dots,x_N\}\in \mathbb{R}^{N\times3}$;\\
        & foreground scores $\mathbf{P}=\{p_1,\dots,p_N\}\in \mathbb{R}^{N}$
    \end{tabular}
    
    \begin{tabular}{l@{}l}
        \textbf{Output: } & sampled key point set $\mathbf{K}=\{k_1,\dots,k_M\}$
    \end{tabular}
    
    \begin{algorithmic}[1]
        \STATE initialize an empty sampling point set $\mathbf{K}$
        \STATE initialize a distance array $d$ of length $N$ with all $+\infty$
        \STATE initialize a visit array $v$ of length $N$ with all zeros
        \FOR{$i=1$ \TO $M$}
            \IF{$i=1$}
                \STATE $k_i = \argmax(\mathbf{P})$
            \ELSE
                \STATE $\mathbf{D} = \{p_k^\gamma \cdot d_k | v_k = 0 \}$
                \STATE $k_i = \argmax(\mathbf{D})$
            \ENDIF
            \STATE add $k_i$ to $\mathbf{K}$, $v_{k_i} = 1$
            \FOR{$j=1$ \TO $N$}
                \STATE $d_j = \min(d_j, \|x_j - x_{k_i}\|)$
            \ENDFOR
        \ENDFOR
        \RETURN $\mathbf{P}$
    \end{algorithmic}
\end{algorithm}

\begin{figure}[t]
    \centering
    \begin{subfigure}[t]{\linewidth}
        \centering
        \includegraphics[width=\linewidth]{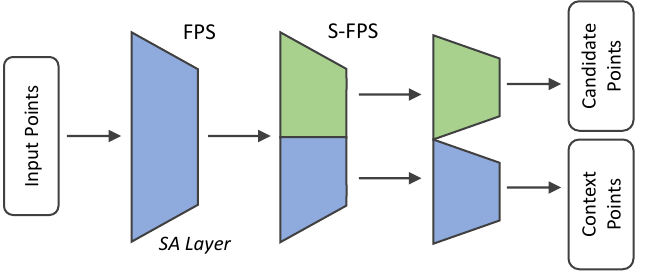}
        \caption{3DSSD backbone}
        \label{fig:3dssd_backbone}
    \end{subfigure}
    
    \begin{subfigure}[t]{\linewidth}
        \centering
        \includegraphics[width=\linewidth]{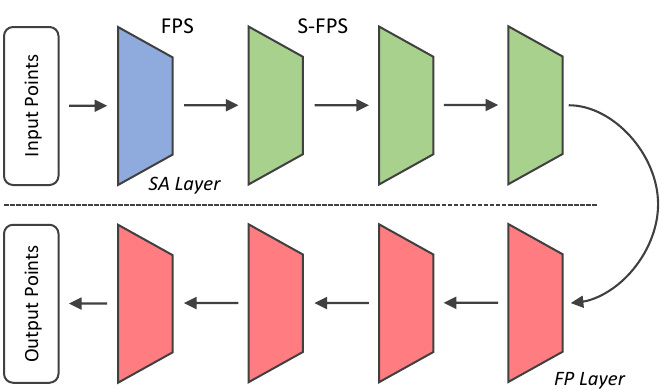}
        \caption{PointRCNN backbone}
        \label{fig:pointrcnn_backbone}
    \end{subfigure}
    \caption{Illustration of 3DSSD and PointRCNN backbones with semantics-augmented set abstraction.}
    \label{fig:backbones}
\end{figure}

\subsection{Implementation Details}\label{sec:implementation_details}

This section provides details about how we implement semantics-augmented set abstraction in 3DSSD \cite{yang20203dssd} and PointRCNN \cite{shi2019pointrcnn}.

\subsubsection{3DSSD.}
\textit{3DSSD} \cite{yang20203dssd} is a lightweight single-stage detector. The backbone is composed of three Pointnet SA layers only and the box prediction network is similar to VoteNet \cite{qi2019deep}, where a vote point indicating the corresponding object centroid is firstly computed from candidate point features and then points in the vicinity of each vote point are aggregated to estimate the 3D box.

3DSSD introduces a \textit{fusion sampling} strategy, where two different point sampling algorithms (namely FPS and F-FPS) are adopted together to sample half of the total key points of the layer respectively. As shown in Figure \ref{fig:3dssd_backbone}, we replace the F-FPS part with our proposed S-FPS and leave all other sampling settings (\textit{e.g.} the number of sampled key points) unchanged. Key points sampled by S-FPS are regarded as candidate points to further generate corresponding vote points and those sampled by FPS are context points for providing context information for nearby votes. We start implementing our SASA from the level 2 SA layer as the raw point input to level 1 cannot produce meaningful semantics. The segmentation loss weights for the level 2 and level 3 SA are set to $0.01$ and $0.1$.

\subsubsection{PointRCNN.}
\textit{PointRCNN} \cite{shi2019pointrcnn} is a representative two-stage detection paradigm with Pointnet. The model comprises a Pointnet++ \cite{qi2017pointnet++} backbone, a point-based RPN and a refinement network. The backbone consists of four SA layers followed by four FP layers. Extracted point features are then fed to the RPN to filter background points and generate 3D regions of interest (RoIs) for foreground points. Finally, the refinement network gathers point features within each RoI and gives the final box estimation.

PointRCNN uses vanilla FPS to sample all key points in all SA layers. As shown in Figure \ref{fig:pointrcnn_backbone}, we apply SASA from level 2 to level 4 and keep the backbone structure, including FP layers, the same as the original implementation. The segmentation loss weights for the three levels are set to $0.001$, $0.01$ and $0.1$.

\section{Experiments}\label{sec:experiments}

\subsection{Datasets}

We validate our semantics-augmented set abstraction on the popular KITTI dataset \cite{geiger2012we} and the more challenging nuScenes dataset \cite{caesar2020nuscenes}.

\subsubsection{KITTI Dataset.}
\textit{KITTI dataset} \cite{geiger2012we} is a prevailing benchmark for 3D object detection in transportation scenarios. It contains $7,481$ LiDAR point clouds as well as finely calibrated 3D bounding boxes for training, and $7,518$ samples for testing. 

Following the commonly applied setting \cite{zhou2018voxelnet}, we divide all training examples into the \textit{train} split ($3,712$ samples) and the \textit{val} split ($3,769$ samples) and all experimental models are trained on the \textit{train} split and tested on the \textit{val} split. For the submission to KITTI \textit{test} server, we follow the training protocol mentioned in \cite{shi2020pv}, where 80\% labeled point cloud images are used for training and the rest 20\% images are used for validation.

\subsubsection{nuScenes Dataset.}
\textit{nuScenes Dataset} \cite{caesar2020nuscenes} is a more challenging dataset for autonomous driving with $380$k LiDAR sweeps from $1,000$ scenes. It is annotated with up to $10$ object categories, including 3D bounding boxes, object velocity and attributes, from the full $360^\circ$ detection range (compared with $90^\circ$ for KITTI). The evaluation metrics used in nuScenes dataset incorporate the commonly used mean average precision (mAP) and a novel nuScenes detection score (NDS), which reflects the overall prediction quality in multiple domains (\textit{i.e.} detection, tracking and attribute estimation).

\subsection{Experiment Settings}

We have two different baselines, 3DSSD \cite{yang20203dssd} and PointRCNN \cite{shi2019pointrcnn}, for evaluation. Our experimental models are all built with the OpenPCDet \cite{openpcdet2020} toolbox, including our reproduced 3DSSD and the official implementation of PointRCNN.

\subsubsection{3DSSD.}
We train the 3DSSD model with ADAM optimizer for $80$ epochs. We apply the one-cycle learning rate schedule \cite{smith2019super} with the peak learning rate at $0.01$. The total batch size is set to $16$, equally distributed on four NVIDIA V100 GPUs.

During the training phase, manifold data augmentation strategies are employed to avoid over-fitting. We use the GT-AUG \cite{yan2018second, shi2019pointrcnn} to paste some instances along with their internal LiDAR points from other scenes to the current training scene. We also utilize global scene augmentations, such as random flipping along the $X$-axis, random scaling with a factor from $[0.9, 1.1]$ and random rotation with an angle from $[-\frac{\pi}{4}, \frac{\pi}{4}]$, as well as box-wise augmentations including random permutation, scaling and rotation. The augmentation settings are kept identical to \cite{yang20203dssd}.  In the inference stage, we use 3D non-maximum-suppression (NMS) with the threshold of $0.01$ to remove redundant predictions.

\begin{table}[t]
\centering
\resizebox{\linewidth}{!}{
\begin{tabular}{c|c c c|c}
\hline
\multirow{2}{*}{Method} & \multicolumn{3}{c|}{Car (IoU=0.7)} & Time\\
& Easy & Mod. & Hard & (ms)\\
\hline
\multicolumn{5}{c}{RGB + LiDAR} \\
\hline
MV3D \cite{chen2017multi} & 74.97 & 63.63 & 54.00 & 360 \\
F-PointNet \cite{qi2018frustum} & 82.19 & 69.79 & 60.59 & 170 \\
AVOD-FPN \cite{ku2018joint} & 83.07 & 71.76 & 65.73 & 100 \\
3D-CVF \cite{yoo20203d} & 89.20 & 80.05 & 73.11 & 75 \\
\hline
\multicolumn{5}{c}{LiDAR only} \\
\hline
\textbf{Voxel-based:} & & & & \\
VoxelNet (Zhou et al. 2018) & 77.47 & 65.11 & 57.73 & 220 \\
SECOND (Yan et al. 2018) & 83.34 & 72.55 & 65.82 & 50 \\
PointPillars \cite{lang2019pointpillars} & 82.58 & 74.31 & 68.99 & \textbf{23.6} \\
TANet \cite{Liu2020TANetR3} & 83.81 & 75.38 & 67.66 & 34.5 \\
Part-$A^2$ \cite{shi2020points} & 87.81 & 78.49 & 73.51 & 80* \\
SA-SSD \cite{he2020structure} & 88.75 & 79.79 & 74.16 & 40.1 \\
CIA-SSD \cite{zheng2020cia} & 89.59 & 80.28 & 72.87 & 30.8 \\
Voxel-RCNN \cite{deng2020voxel} & \textbf{90.90} & 81.62 & 77.06 & 25.2 \\
\hline
\textbf{PV-based:} & & & & \\
F-PointRCNN \cite{chen2019fast} & 84.28 & 75.73 & 67.39 & 65 \\
PV-RCNN \cite{shi2020pv} & 90.25 & 81.43 & 76.82 & 80* \\
\hline
\textbf{Point-based:} & & & & \\
PointRCNN (Shi et al. 2019) & 86.96 & 75.64 & 70.70 & 100* \\
STD \cite{yang2019std} & 87.95 & 79.71 & 75.09 & 80 \\
3DSSD \cite{yang20203dssd} & 88.36 & 79.57 & 74.55 & 38 \\
Ours (3DSSD + \textbf{SASA}) & 88.76 & \textbf{82.16} & \textbf{77.16} & 36 \\
\hline
\end{tabular}
}
\caption{Results on the car class of KITTI \textit{test} set. Our model is 3DSSD with SASA. The evaluation metric is the AP calculated on $40$ recall points. Inference time data with ``*" is pasted from the official KITTI benchmark website.}
\label{tab: KITTI}
\end{table}

\begin{table*}[t]
\centering
\resizebox{\linewidth}{!}{
\begin{tabular}{c|c|c||c|c|c|c|c|c|c|c|c|c}
\hline
Method & NDS & mAP & Car & Truck & Bus & Trailer & C.V. & Ped. & Motor & Bicycle & T.C. & Barrier \\
\hline
PointPillars \cite{lang2019pointpillars} & 46.8 & 28.2 & 75.5 & 31.6 & 44.9 & 23.7 & 4.0 & 49.6 & 14.6 & 0.4 & 8.0 & 30.0 \\
3D-CVF \cite{yoo20203d} & 49.8 & 42.2 & 79.7 & 37.9 & 55.0 & 36.3 & - & 71.3 & 37.2 & - & 40.8 & 47.1 \\
3DSSD \cite{yang20203dssd} & 56.4 & 42.6 & 81.2 & 47.2 & 61.4 & 30.5 & 12.6 & 70.2 & 36.0 & 8.6 & 31.1 & 47.9 \\
Ours (3DSSD + \textbf{SASA}) & \textbf{61.0} & \textbf{45.0} & 76.8 & 45.0 & 66.2 & 36.5 & 16.1 & 69.1 & 39.6 & 16.9 & 29.9 & 53.6 \\
\hline
\end{tabular}}
\caption{Results on the nuScenes \textit{validation} set. Our model is 3DSSD with SASA. Evaluation metrics include NDS, mAP and AP on $10$ classes. Abbreviations: Pedestrian (Ped.), Traffic cone (T.C.), Construction vehicle (C.V.).}
\label{tab: nuScenes}
\end{table*}

\medskip

When transferring to the nuScenes dataset, we follow the official suggestion \cite{caesar2020nuscenes} that combining LiDAR points from the current key frame as well as previous frames in $0.5$s, which involves up to $400$k LiDAR points in a single training sample. Then, we reduce the quantity of input LiDAR points in the same way as \cite{yang20203dssd}. In particular, we voxelize the point cloud from the key frame as well as that from piled previous frames with the voxel size of $(0.1$m$, 0.1$m$, 0.1$m$)$, then randomly select $16,384$ and $49,152$ voxels from the key frame and previous frames and randomly choose one internal LiDAR point from each selected voxel. The total $65,536$ LiDAR points with 3D coordinates, reflectance and timestamp \cite{caesar2020nuscenes} are fed to the model. The training phase runs for $20$ epochs with a batch size of $16$ on eight NVIDIA V100 GPUs.

\subsubsection{PointRCNN.}
According to the model configuration provided in OpenPCDet \cite{openpcdet2020}, we train PointRCNN \cite{shi2019pointrcnn} in an end-to-end manner with ADAM optimizer for $80$ epochs. The learning rate schedule is one-cycle schedule \cite{smith2019super} with a peak learning rate at $0.01$. We follow the original data augmentation strategies and inference settings. Please refer to \cite{shi2019pointrcnn} and \cite{openpcdet2020} for more details.

\subsection{Main Results.}
Our main evaluation compared with state-of-the-art models is performed on the 3DSSD model with our proposed SASA.

\subsubsection{Results on KITTI Dataset.}
Table \ref{tab: KITTI} shows the 3D object detection performance on the KITTI \textit{test} set evaluated on the official server. For the most competitive car detection race track, our method surpasses all existing point-based detectors by a great margin and obtains comparable results to state-of-the-art voxel-based models. Comparing with the baseline model 3DSSD, our method boosts the AP by $0.40\%, 2.59\%, 2.61\%$ for the three difficulty levels respectively. It is worth noting that our method acquires significant improvements on the moderate and hard levels, demonstrating our proposed semantics-augmented operation can retain sufficient points from hardly visible instances so as to make more robust object estimations, which is of great significance in building safe autonomous driving systems.

\subsubsection{Results on nuScenes Dataset.}
We report the nuScenes detection score (NDS) and the mean average precision (mAP) as well as the average precision (AP) for the $10$ object categories in Table \ref{tab: nuScenes}. Our method obtains much higher NDS and mAP compared with the baseline method 3DSSD ($4.6\%$ on NDS and $2.4\%$ on mAP). We believe our proposed SASA efficiently chooses plenty of key points from multiple frames so as to enhance the detection accuracy as well as the tracking accuracy. Especially for bicycles that are commonly regarded as difficult detection targets, our method still produces satisfactory results.

\subsubsection{Inference Speed.}
Our model takes around $36$ms to process a single point cloud sample from KITTI dataset, measured with OpenPCDet \cite{openpcdet2020} framework on a V100 GPU. Compared with F-FPS, S-FPS bypasses the time-consuming calculation of the pairwise feature distance. Especially when the number of points becomes large, our strategy avoids quadratic growth of computations and GPU memory usage for sampling.

\begin{figure*}[t]
\centering
\includegraphics[width=0.95\linewidth]{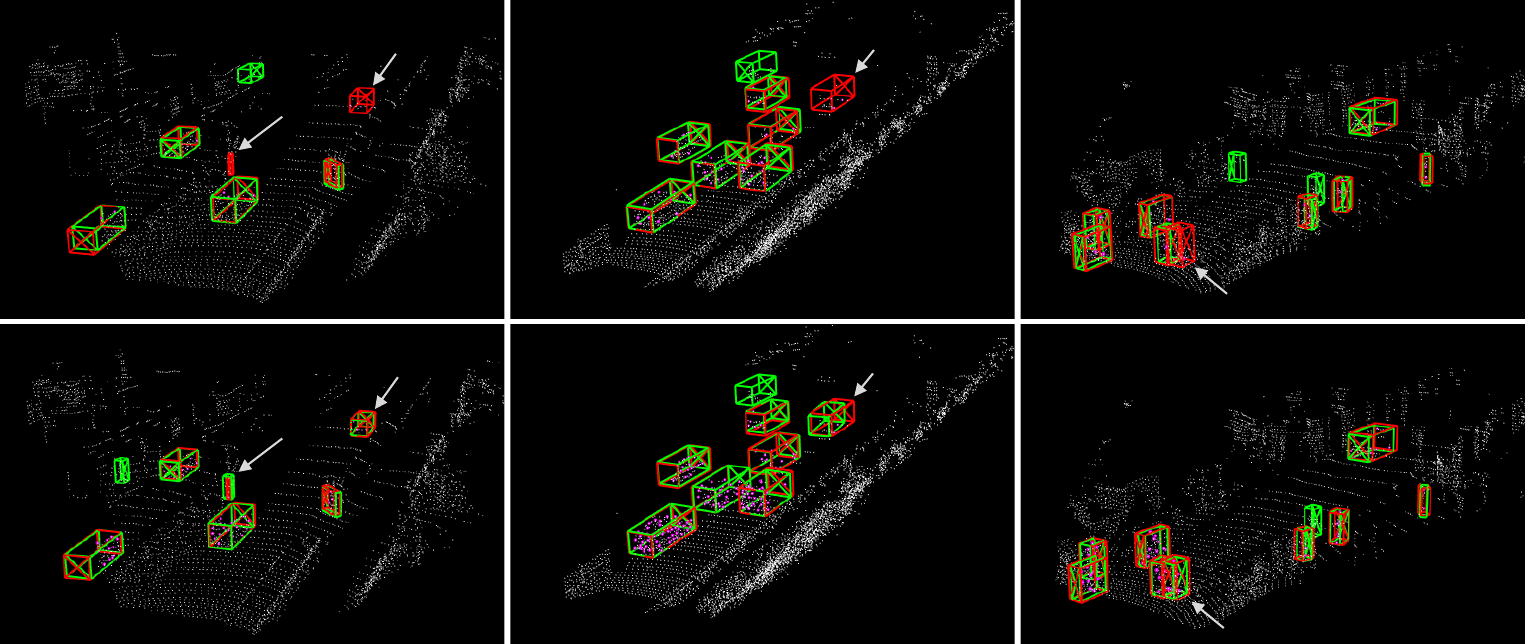}
\caption{Visualizing detection results between F-FPS (\textit{top}) and S-FPS (\textit{bottom}) on the KITTI \textit{val} split. Predicted and ground-truth boxes are labeled in green and red, respectively. Pink points mark the key points sampled in the last SA layer and white arrows indicate some false negative examples for F-FPS but successfully recovered by our method.}
\label{fig:visulization}
\end{figure*}

\subsection{Ablation Study}

We conduct ablation studies to validate each part of SASA. All results provided in this section are trained on the KITTI \textit{train} split and evaluated on the \textit{val} split of the car class.

\begin{table}[t]
\centering
\resizebox{\linewidth}{!}{
\begin{tabular}{l | >{\centering\arraybackslash}m{0.25cm} >{\centering\arraybackslash}m{0.25cm} | c c c | c}
\hline
Sampling Method & PS & FS & Easy & Mod. & Hard & Recall \\
\hline
FPS & \xmark & \xmark & 91.08 & 82.75 & 79.93 & 92.10 \\ %
FPS & \cmark & \xmark & 91.17 & 82.83 & 81.97 & 92.01 \\
\hline
F-FPS & \xmark & \cmark & 91.54 & 83.46 & 82.18 & 96.65 \\
S-FPS ($\gamma=0.1$) & \cmark & \cmark & 91.53 & 83.16 & 81.92 & 95.79 \\
S-FPS ($\gamma=1$) & \cmark & \cmark & \textbf{92.19} & \textbf{85.76} & \textbf{83.11} & \textbf{97.65} \\
S-FPS ($\gamma=10$) & \cmark & \cmark & 92.17 & 83.41 & 80.61 & 95.02  \\
S-FPS ($\gamma=100$) & \cmark & \cmark & 91.72 & 82.35 & 78.24 & 91.19 \\
\hline
\end{tabular}}
\caption{Performance comparison between FPS, F-FPS and S-FPS with different balance factor settings. ``PS" represents point segmentation modules and ``FS" represents the fusion sampling strategy. Point recall is calculated according to the $256$ candidate points that are used to generate votes.}
\label{tab: ablation_sampling}
\end{table}

\subsubsection{Effects of Semantics-guided Point Sampling.}
Table \ref{tab: ablation_sampling} compares the detection performance as well as the point recall, which means that the proportion of GT boxes that have at least one internal sample point comparing to the total number of GT boxes \cite{yang20203dssd}, among different sampling algorithms, based on the 3DSSD baseline. We only adjust the point sampling strategy and keep other model settings identical. Results show that our S-FPS outperforms the F-FPS used in the 3DSSD baseline in all three difficulty levels, especially by up to $2.30\%$ in the moderate level. Also, candidate points sampled by our method can ``hit" $1\%$ more ground-truth boxes comparing to F-FPS.

Visualization results in Figure \ref{fig:visulization} also prove our method effective. Comparing to F-FPS, S-FPS can keep more key points within a single instance, even for those severely occluded or tiny objects. Thus, hard examples are more likely to be detected with our proposed S-FPS sampling algorithm.

\subsubsection{Effects of Point Segmentation Layer.}
The $1^{st}$ row and the $2^{nd}$ row of Table \ref{tab: ablation_sampling} compare the detection performance with and without point segmentation modules. Stand-alone segmentation layers show limited effects on the detection accuracy. The improvement is mainly derived from the point sampling algorithm.

\subsubsection{Effects of Semantics Balance Factor.}
We also compare S-FPS with different balance factor $\gamma$ from the $4^{th}$ to $7^{th}$ row in Table \ref{tab: ablation_sampling}. Results indicate that an overly large or small $\gamma$ could not appropriately boost the detection accuracy. As aforementioned, S-FPS will approximate to the top-K selection on foreground scores if the $\gamma$ becomes extremely large. Sampled key points could crowd in a minority of easily identified instances and fail to cover distant or occluded ones. When $\gamma=100$, the point recall drops sharply to $91.19\%$, even worse than that for vanilla FPS. Also, the box prediction network would encounter the imbalance training problem as the quantity of internal sampled points shows a great disparity among objects. Therefore, the overall detection performance is largely hurt. From the other aspect, S-FPS will degrade to vanilla FPS if $\gamma$ is close to $0$, making limited improvement. A suitable $\gamma$ could significantly improve the performance. When $\gamma=1$, the three difficulty levels reach satisfactory performance simultaneously.

\subsection{Compatibility Study}

\begin{table}[t]
\centering \small
\begin{tabular}{c | c c c | c}
\hline
Method & Easy & Mod. & Hard & mAP \\
\hline
3DSSD & 91.54 & 83.46 & 82.18 & 85.73 \\
3DSSD + \textbf{SASA} & 92.19 & 85.76 & 83.11 & 87.02 \\
\rowcolor{LightCyan}
\textit{SASA Improvement} & \textit{+0.65} & \textit{+2.30} & \textit{+0.93} & \textit{+1.29} \\
\hline
PointRCNN & 91.57 & 82.24 & 80.45 & 84.75\\
PointRCNN + \textbf{SASA} & 92.13 & 82.64 & 82.40 & 85.72 \\
\rowcolor{LightCyan}
\textit{SASA Improvement} & \textit{+0.56} & \textit{+0.40} & \textit{+1.95} & \textit{+0.97} \\
\hline
\end{tabular}
\caption{Effects of SASA in different point-based detection paradigms evaluated on the car class of KITTI \textit{val} split.}
\label{tab: ablation_pointrcnn}
\end{table}

\begin{figure}[t]
    \centering
    \includegraphics[width=0.95\linewidth]{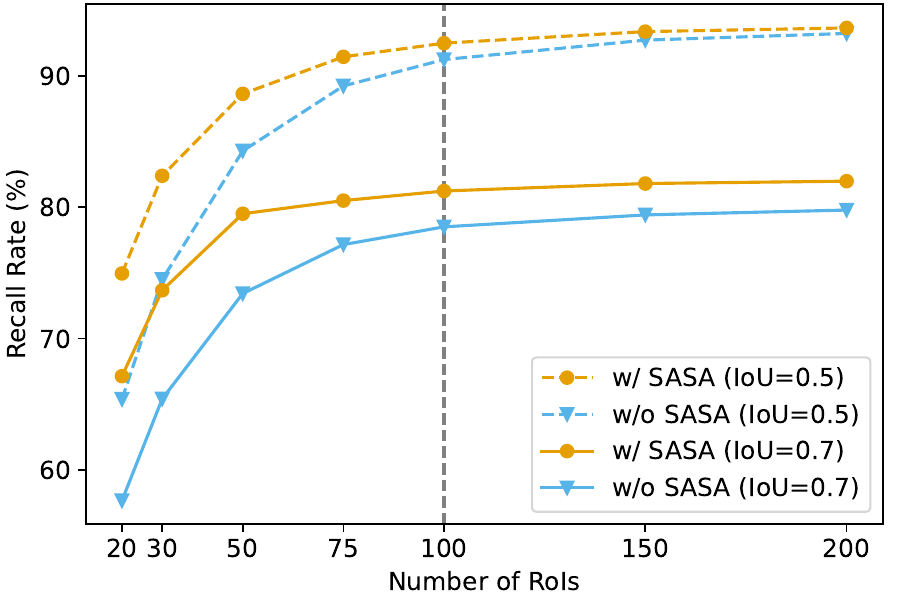}
    \caption{Proposal recall rate under different numbers of RoIs with and without SASA on PointRCNN.}
    \label{fig:ablation_proposal}
\end{figure}

Our SASA is an easy-to-plug-in design and can serve multiple point-based detection paradigms. As SASA already obtains notable enhancement on the one-stage model 3DSSD (as shown in Table \ref{tab: ablation_pointrcnn}), here we test its compatibility in a two-stage point-based detector, PointRCNN.

\subsubsection{Results on KITTI Dataset.}
Although the PointRCNN backbone contains four FP layers to recover points discarded in the SA stage, there is no concern about the point recall rate. Nevertheless, SASA can still enhance the detection performance. As shown in Table \ref{tab: ablation_pointrcnn}, it improves the detection performance for all difficulty levels, especially with a $1.95\%$ AP leading in the hard mode. 
Harder instances hold fewer LiDAR points, which are usually difficult to survive in deep SA layers with vanilla FPS. Their features could not be deeply and sufficiently encoded by the Pointnet backbone, while S-FPS can better focus on these point samples and their feature quality is largely improved. 

\subsubsection{Quantitative Analysis.}
Here we further analyze the quality of produced point features by comparing the accuracy of 3D proposals generated by RPN. From Figure \ref{fig:ablation_proposal}, point features extracted by our semantics-augmented backbone can yield more accurate 3D RoIs compared to the original one. When we use the top-$100$ RoIs as PointRCNN suggests, our method can cover nearly $2\%$ more ground-truth boxes at the IoU level of $0.7$. This gap widens to almost $10\%$ when the number of RoIs becomes lower, demonstrating the superiority of the our feature learning scheme with SASA.

\section{Conclusion}

In this paper, we present the Semantics-Augmented Set Abstraction (SASA) for point-based 3D detection. Our main concept is to incorporate semantic information into the Pointnet SA stage for guiding the point-based backbone to better model potential objects. Experimental results on the KITTI and nuScenes datasets indicate that our strategy can help access a higher point recall during the point down-sampling stage so as to obtain a better detection outcome for manifold point-based detectors. Our proposed method provides a promising direction for point-based detection. Not only can it be implemented in Pointnet-based models, but it is also compatible with, for example, transformer-based networks and graph neural networks for model reduction. We hope this study could inspire the research community to further break the sampling bottleneck in point-based detection.

\bibliography{aaai22}

\begin{thebibliography}{32}
\providecommand{\natexlab}[1]{#1}

\bibitem[{Beltr{\'a}n et~al.(2018)Beltr{\'a}n, Guindel, Moreno, Cruzado,
  Garcia, and De~La~Escalera}]{beltran2018birdnet}
Beltr{\'a}n, J.; Guindel, C.; Moreno, F.~M.; Cruzado, D.; Garcia, F.; and
  De~La~Escalera, A. 2018.
\newblock Birdnet: a 3d object detection framework from lidar information.
\newblock In \emph{2018 21st International Conference on Intelligent
  Transportation Systems (ITSC)}, 3517--3523. IEEE.

\bibitem[{Caesar et~al.(2020)Caesar, Bankiti, Lang, Vora, Liong, Xu, Krishnan,
  Pan, Baldan, and Beijbom}]{caesar2020nuscenes}
Caesar, H.; Bankiti, V.; Lang, A.~H.; Vora, S.; Liong, V.~E.; Xu, Q.; Krishnan,
  A.; Pan, Y.; Baldan, G.; and Beijbom, O. 2020.
\newblock nuscenes: A multimodal dataset for autonomous driving.
\newblock In \emph{Proceedings of the IEEE/CVF conference on computer vision
  and pattern recognition}, 11621--11631.

\bibitem[{Chen et~al.(2017)Chen, Ma, Wan, Li, and Xia}]{chen2017multi}
Chen, X.; Ma, H.; Wan, J.; Li, B.; and Xia, T. 2017.
\newblock Multi-view 3d object detection network for autonomous driving.
\newblock In \emph{Proceedings of the IEEE Conference on Computer Vision and
  Pattern Recognition}, 1907--1915.

\bibitem[{Chen et~al.(2019)Chen, Liu, Shen, and Jia}]{chen2019fast}
Chen, Y.; Liu, S.; Shen, X.; and Jia, J. 2019.
\newblock Fast point r-cnn.
\newblock In \emph{Proceedings of the IEEE/CVF International Conference on
  Computer Vision}, 9775--9784.

\bibitem[{Chen, Zhang, and Tao(2019)}]{chen2019progressive}
Chen, Z.; Zhang, J.; and Tao, D. 2019.
\newblock Progressive lidar adaptation for road detection.
\newblock \emph{IEEE/CAA Journal of Automatica Sinica}, 6(3): 693--702.

\bibitem[{Deng et~al.(2020)Deng, Shi, Li, Zhou, Zhang, and Li}]{deng2020voxel}
Deng, J.; Shi, S.; Li, P.; Zhou, W.; Zhang, Y.; and Li, H. 2020.
\newblock Voxel R-CNN: Towards High Performance Voxel-based 3D Object
  Detection.
\newblock \emph{arXiv preprint arXiv:2012.15712}.

\bibitem[{Geiger, Lenz, and Urtasun(2012)}]{geiger2012we}
Geiger, A.; Lenz, P.; and Urtasun, R. 2012.
\newblock Are we ready for autonomous driving? the kitti vision benchmark
  suite.
\newblock In \emph{2012 IEEE Conference on Computer Vision and Pattern
  Recognition}, 3354--3361. IEEE.

\bibitem[{He et~al.(2020)He, Zeng, Huang, Hua, and Zhang}]{he2020structure}
He, C.; Zeng, H.; Huang, J.; Hua, X.-S.; and Zhang, L. 2020.
\newblock Structure aware single-stage 3d object detection from point cloud.
\newblock In \emph{Proceedings of the IEEE/CVF Conference on Computer Vision
  and Pattern Recognition}, 11873--11882.

\bibitem[{Ku et~al.(2018)Ku, Mozifian, Lee, Harakeh, and
  Waslander}]{ku2018joint}
Ku, J.; Mozifian, M.; Lee, J.; Harakeh, A.; and Waslander, S.~L. 2018.
\newblock Joint 3d proposal generation and object detection from view
  aggregation.
\newblock In \emph{2018 IEEE/RSJ International Conference on Intelligent Robots
  and Systems (IROS)}, 1--8. IEEE.

\bibitem[{Lang et~al.(2019)Lang, Vora, Caesar, Zhou, Yang, and
  Beijbom}]{lang2019pointpillars}
Lang, A.~H.; Vora, S.; Caesar, H.; Zhou, L.; Yang, J.; and Beijbom, O. 2019.
\newblock Pointpillars: Fast encoders for object detection from point clouds.
\newblock In \emph{Proceedings of the IEEE/CVF Conference on Computer Vision
  and Pattern Recognition}, 12697--12705.

\bibitem[{Lang, Manor, and Avidan(2020)}]{lang2020samplenet}
Lang, I.; Manor, A.; and Avidan, S. 2020.
\newblock SampleNet: differentiable point cloud sampling.
\newblock In \emph{Proceedings of the IEEE/CVF Conference on Computer Vision
  and Pattern Recognition}, 7578--7588.

\bibitem[{Liu et~al.(2015)Liu, Wang, Foroosh, Tappen, and
  Pensky}]{liu2015sparse}
Liu, B.; Wang, M.; Foroosh, H.; Tappen, M.; and Pensky, M. 2015.
\newblock Sparse convolutional neural networks.
\newblock In \emph{Proceedings of the IEEE conference on computer vision and
  pattern recognition}, 806--814.

\bibitem[{Liu et~al.(2020)Liu, Zhao, Huang, Hu, Zhou, and Bai}]{Liu2020TANetR3}
Liu, Z.; Zhao, X.; Huang, T.; Hu, R.; Zhou, Y.; and Bai, X. 2020.
\newblock TANet: Robust 3D Object Detection from Point Clouds with Triple
  Attention.
\newblock \emph{ArXiv}, abs/1912.05163.

\bibitem[{Nezhadarya et~al.(2020)Nezhadarya, Taghavi, Razani, Liu, and
  Luo}]{nezhadarya2020adaptive}
Nezhadarya, E.; Taghavi, E.; Razani, R.; Liu, B.; and Luo, J. 2020.
\newblock Adaptive hierarchical down-sampling for point cloud classification.
\newblock In \emph{Proceedings of the IEEE/CVF Conference on Computer Vision
  and Pattern Recognition}, 12956--12964.

\bibitem[{Qi et~al.(2019)Qi, Litany, He, and Guibas}]{qi2019deep}
Qi, C.~R.; Litany, O.; He, K.; and Guibas, L.~J. 2019.
\newblock Deep hough voting for 3d object detection in point clouds.
\newblock In \emph{Proceedings of the IEEE International Conference on Computer
  Vision}, 9277--9286.

\bibitem[{Qi et~al.(2018)Qi, Liu, Wu, Su, and Guibas}]{qi2018frustum}
Qi, C.~R.; Liu, W.; Wu, C.; Su, H.; and Guibas, L.~J. 2018.
\newblock Frustum pointnets for 3d object detection from rgb-d data.
\newblock In \emph{Proceedings of the IEEE conference on computer vision and
  pattern recognition}, 918--927.

\bibitem[{Qi et~al.(2017{\natexlab{a}})Qi, Su, Mo, and Guibas}]{qi2017pointnet}
Qi, C.~R.; Su, H.; Mo, K.; and Guibas, L.~J. 2017{\natexlab{a}}.
\newblock Pointnet: Deep learning on point sets for 3d classification and
  segmentation.
\newblock In \emph{Proceedings of the IEEE conference on computer vision and
  pattern recognition}, 652--660.

\bibitem[{Qi et~al.(2017{\natexlab{b}})Qi, Yi, Su, and
  Guibas}]{qi2017pointnet++}
Qi, C.~R.; Yi, L.; Su, H.; and Guibas, L.~J. 2017{\natexlab{b}}.
\newblock Pointnet++: Deep hierarchical feature learning on point sets in a
  metric space.
\newblock In \emph{Advances in neural information processing systems},
  5099--5108.

\bibitem[{Shi et~al.(2020{\natexlab{a}})Shi, Guo, Jiang, Wang, Shi, Wang, and
  Li}]{shi2020pv}
Shi, S.; Guo, C.; Jiang, L.; Wang, Z.; Shi, J.; Wang, X.; and Li, H.
  2020{\natexlab{a}}.
\newblock Pv-rcnn: Point-voxel feature set abstraction for 3d object detection.
\newblock In \emph{Proceedings of the IEEE/CVF Conference on Computer Vision
  and Pattern Recognition}, 10529--10538.

\bibitem[{Shi, Wang, and Li(2019)}]{shi2019pointrcnn}
Shi, S.; Wang, X.; and Li, H. 2019.
\newblock Pointrcnn: 3d object proposal generation and detection from point
  cloud.
\newblock In \emph{Proceedings of the IEEE/CVF Conference on Computer Vision
  and Pattern Recognition}, 770--779.

\bibitem[{Shi et~al.(2020{\natexlab{b}})Shi, Wang, Shi, Wang, and
  Li}]{shi2020points}
Shi, S.; Wang, Z.; Shi, J.; Wang, X.; and Li, H. 2020{\natexlab{b}}.
\newblock From points to parts: 3d object detection from point cloud with
  part-aware and part-aggregation network.
\newblock \emph{IEEE transactions on pattern analysis and machine
  intelligence}.

\bibitem[{Smith and Topin(2019)}]{smith2019super}
Smith, L.~N.; and Topin, N. 2019.
\newblock Super-convergence: Very fast training of neural networks using large
  learning rates.
\newblock In \emph{Artificial Intelligence and Machine Learning for
  Multi-Domain Operations Applications}, volume 11006, 1100612. International
  Society for Optics and Photonics.

\bibitem[{Song and Xiao(2016)}]{song2016deep}
Song, S.; and Xiao, J. 2016.
\newblock Deep sliding shapes for amodal 3d object detection in rgb-d images.
\newblock In \emph{Proceedings of the IEEE conference on computer vision and
  pattern recognition}, 808--816.

\bibitem[{Team(2020)}]{openpcdet2020}
Team, O.~D. 2020.
\newblock OpenPCDet: An Open-source Toolbox for 3D Object Detection from Point
  Clouds.
\newblock \url{https://github.com/open-mmlab/OpenPCDet}.

\bibitem[{Yan, Mao, and Li(2018)}]{yan2018second}
Yan, Y.; Mao, Y.; and Li, B. 2018.
\newblock Second: Sparsely embedded convolutional detection.
\newblock \emph{Sensors}, 18(10): 3337.

\bibitem[{Yang et~al.(2019{\natexlab{a}})Yang, Zhang, Ni, Li, Liu, Zhou, and
  Tian}]{yang2019modeling}
Yang, J.; Zhang, Q.; Ni, B.; Li, L.; Liu, J.; Zhou, M.; and Tian, Q.
  2019{\natexlab{a}}.
\newblock Modeling point clouds with self-attention and gumbel subset sampling.
\newblock In \emph{Proceedings of the IEEE/CVF Conference on Computer Vision
  and Pattern Recognition}, 3323--3332.

\bibitem[{Yang et~al.(2020)Yang, Sun, Liu, and Jia}]{yang20203dssd}
Yang, Z.; Sun, Y.; Liu, S.; and Jia, J. 2020.
\newblock 3dssd: Point-based 3d single stage object detector.
\newblock In \emph{Proceedings of the IEEE/CVF Conference on Computer Vision
  and Pattern Recognition}, 11040--11048.

\bibitem[{Yang et~al.(2019{\natexlab{b}})Yang, Sun, Liu, Shen, and
  Jia}]{yang2019std}
Yang, Z.; Sun, Y.; Liu, S.; Shen, X.; and Jia, J. 2019{\natexlab{b}}.
\newblock Std: Sparse-to-dense 3d object detector for point cloud.
\newblock In \emph{Proceedings of the IEEE/CVF International Conference on
  Computer Vision}, 1951--1960.

\bibitem[{Yoo et~al.(2020)Yoo, Kim, Kim, and Choi}]{yoo20203d}
Yoo, J.~H.; Kim, Y.; Kim, J.~S.; and Choi, J.~W. 2020.
\newblock 3d-cvf: Generating joint camera and lidar features using cross-view
  spatial feature fusion for 3d object detection.
\newblock \emph{arXiv preprint arXiv:2004.12636}, 3.

\bibitem[{Zhang and Tao(2020)}]{zhang2020empowering}
Zhang, J.; and Tao, D. 2020.
\newblock Empowering things with intelligence: a survey of the progress,
  challenges, and opportunities in artificial intelligence of things.
\newblock \emph{IEEE Internet of Things Journal}, 8(10): 7789--7817.

\bibitem[{Zheng et~al.(2020)Zheng, Tang, Chen, Jiang, and Fu}]{zheng2020cia}
Zheng, W.; Tang, W.; Chen, S.; Jiang, L.; and Fu, C.-W. 2020.
\newblock CIA-SSD: Confident IoU-Aware Single-Stage Object Detector From Point
  Cloud.
\newblock \emph{arXiv preprint arXiv:2012.03015}.

\bibitem[{Zhou and Tuzel(2018)}]{zhou2018voxelnet}
Zhou, Y.; and Tuzel, O. 2018.
\newblock Voxelnet: End-to-end learning for point cloud based 3d object
  detection.
\newblock In \emph{Proceedings of the IEEE Conference on Computer Vision and
  Pattern Recognition}, 4490--4499.

\end{thebibliography}

\clearpage

\appendix

\section{Further Analysis of Point Sampling}
In this section, we present more detailed analysis of point sampling methods by layer on 3DSSD \cite{yang20203dssd} and PointRCNN \cite{shi2019pointrcnn}. We evaluate the sampling performance with two metrics, the aforementioned point recall rate and the foreground rate, which means the proportion of foreground key points compared to the total number of sampled key points in each layer.

\subsection{Analysis of Sampling Performance}
Here we present a thorough study on point sampling performance by layer. The results for 3DSSD are given in Table \ref{tab: recall_3dssd}. Although S-FPS does not achieve noticeable point recall leading compared with F-FPS, we can see much difference from the foreground rate. In the level 2 and level 3 SA layers, F-FPS can only maintain a foreground rate around $9\%$. More than $90\%$ key points are sampled from background, running the risk of delivering trivial information for detection. While S-FPS can enhance the foreground rate to $35.23\%$ and $31.24\%$ for the level 2 and level 3 SA respectively. This proves that our proposed SASA can provide richer foreground point representations for the succeeding box prediction network so as to further improve the detection performance.

SASA is even more effective on PointRCNN. As shown in Table \ref{tab: recall_pointrcnn}, in the $4^{th}$ SA layer, the vanilla FPS could only cover $39.62\%$ ground-truth instances and only $2.4\%$ of sample points are from foreground. A large quantity of foreground instances are somehow ignored in the level 4 SA and their related point representations cannot be processed sufficiently. While our proposed S-FPS in SASA can still reach a very high recall rate at $96.36$ with only $64$ key points and $31.2\%$ of sampled key points are foreground points.

\subsection{Analysis of the Level 2 SA Layer}
Point semantic map in SASA is calculated with the point features given by the preceding SA layer. Since the receptive field of the level 1 SA is relatively limited, the point segmentation module in the level 2 SA may not be able to produce accurate point semantic map with point features from the level 1 SA. It is concerning that whether the S-FPS in the level 2 SA could function well with less accurate point-wise semantics. As shown in Table \ref{tab: recall_3dssd} and Table \ref{tab: recall_pointrcnn}, the results demonstrate that implementing SASA from level 2 is still beneficial. The S-FPS for the level 2 SA still brings some improvements on both metrics compared with baseline sampling methods.

\begin{table*}[t]
\centering
\begin{tabular}{l | l | c | c c | c c }
\hline
\multirow{2}{*}{Method} & SA Layer & Level 1 & \multicolumn{2}{c|}{Level 2} & \multicolumn{2}{c}{Level 3} \\
& Total Point \# & 4096 & 512 & 512 & 256 & 256 \\
\hline
\multirow{3}{*}{3DSSD} & Sampling Method & FPS & F-FPS & FPS & F-FPS & FPS  \\
& Foreground Rate & 4.40 & 9.09 & 3.20 & 8.70 & 2.75 \\
& Point Recall & 98.35 & 97.75 & 96.73 & 96.65 & 91.58 \\
\hline
\multirow{3}{*}{3DSSD + \textbf{SASA}} & Sampling Method & FPS & S-FPS & FPS & S-FPS & FPS \\
& Foreground Rate & 4.40 & \textbf{35.23} & 3.20 & \textbf{31.24} & 2.75 \\
& Point Recall & 98.35 & \textbf{97.87} & 96.73 & \textbf{97.65} & 91.58 \\
\hline
\rowcolor{LightCyan}
& Foreground Rate & - & \textit{+26.14} & - & \textit{+22.54} & -\\
\rowcolor{LightCyan}
\multirow{-2}{*}{\textit{SASA Improvement}} & Point Recall & \textit{-} & \textit{+0.12} & \textit{-} & \textit{+1.00} & - \\
\hline
\end{tabular}
\caption{Analysis of point sampling by layer on 3DSSD, evaluated on the car class of KITTI \textit{val} split. The level 2 and level 3 SA layer exploit the fusion sampling strategy to sample half of key points with two different sampling algorithms individually.}
\label{tab: recall_3dssd}
\end{table*}

\begin{table*}[t]
\centering
\begin{tabular}{l | l  c c c c}
\hline
\multirow{2}{*}{Method} & SA Layer & Level 1 & Level 2 & Level 3 & Level 4 \\
& Total Point \# & 4096 & 1024 & 256 & 64 \\
\hline
\multirow{3}{*}{PointRCNN} & Sampling Method & FPS & FPS & FPS & FPS \\
& Foreground Rate & 4.41 & 3.55 & 2.74 & 2.41 \\
& Point Recall & 98.35 & 97.91 & 91.81 & 39.62 \\
\hline
\multirow{3}{*}{PointRCNN + \textbf{SASA}} & Sampling Method & FPS & S-FPS & S-FPS & S-FPS \\
& Foreground Rate & 4.41 & \textbf{15.11} & \textbf{28.35} & \textbf{37.02} \\
& Point Recall & 98.35 & \textbf{98.18} & \textbf{97.73} & \textbf{96.36} \\
\hline
\rowcolor{LightCyan}
& Foreground Rate & - & \textit{+11.56} & \textit{+25.61} & \textit{+34.61} \\
\rowcolor{LightCyan}
\multirow{-2}{*}{\textit{SASA Improvement}} & Point Recall & \textit{-} & \textit{+0.27} & \textit{+5.92} & \textit{+56.74} \\
\hline
\end{tabular}
\caption{Analysis of point sampling by layer on PointRCNN, evaluated on the car class of KITTI \textit{val} split.}
\label{tab: recall_pointrcnn}
\end{table*}

\end{document}